\documentclass{article}
\usepackage{ijcai24}

\usepackage{times}
\usepackage{soul}
\usepackage{url}
\usepackage[hidelinks]{hyperref}
\usepackage[utf8]{inputenc}
\usepackage[small]{caption}
\usepackage{graphicx}
\usepackage{amsmath}
\usepackage{amsfonts}
\usepackage{amsthm}
\usepackage{booktabs}
\usepackage{algorithm}
\usepackage{algorithmic}
\usepackage[switch]{lineno}
\usepackage{balance}
\usepackage{float}
\usepackage{xcolor}
\usepackage{multirow} 
\usepackage{makecell}
\usepackage{enumitem}
\usepackage{color, colortbl}
\definecolor{Gray}{gray}{0.9}
\usepackage{amssymb}
\usepackage{enumitem}

\newtheorem{problem}{Problem Definition}

\title{Beyond Generalization: A Survey of Out-Of-Distribution Adaptation on Graphs}

\author{
Shuhan Liu$^1$
\And
Kaize Ding$^1$
\affiliations
$^1$Northwestern University, Evanston, IL, USA
\emails
shuhanliu@u.northwestern.edu, kaize.ding@northwestern.edu
}

\begin{document}

\maketitle

\begin{abstract}
Distribution shifts on graphs -- the data distribution discrepancies between training and testing a graph machine learning model, are often ubiquitous and unavoidable in real-world scenarios. Such shifts may severely deteriorate the performance of the model, posing significant challenges for reliable graph machine learning. Consequently, there has been a surge in research on graph Out-Of-Distribution (OOD) adaptation methods that aim to mitigate the distribution shifts and adapt the knowledge from one distribution to another. In our survey, we provide an up-to-date and forward-looking review of graph OOD adaptation methods, covering two main problem scenarios including training-time as well as test-time graph OOD adaptation.  We start by formally formulating the two problems and then discuss different types of distribution shifts on graphs.
Based on our proposed taxonomy for graph OOD adaptation, we systematically categorize the existing methods according to their learning paradigm and investigate the techniques behind them. 
Finally, we point out promising research directions and the
corresponding challenges. We also provide a continuously updated reading list at \url{https://github.com/kaize0409/Awesome-Graph-OOD-Adaptation.git}

\end{abstract}

\section{Introduction}
\label{sec:introduction}
Motivated by the prevalence of graph-structured data in various real-world scenarios, growing attention has been paid to graph machine learning, which seeks to efficiently capture relationships and dependencies among entities within graphs.
In particular, Graph Neural Networks (GNNs) are able to effectively learn the representations on graphs through message-passing ~\cite{Kipf2016Semi,wu2019simplifying,hamilton2017inductive}, which have demonstrated remarkable success across diverse applications, such as social networks, physics problems, and traffic networks~\cite{Bi2023PredictingTS,Liu2023StructuralRI,Zhu2020TransferLO}.


While graph machine learning has achieved notable success, most of the existing efforts presume that test data follows the same distribution as training data, which is often invalid in the wild.
The performance of traditional graph machine learning methods may substantially degrade when confronted with Out-Of-Distribution (OOD) samples, limiting their efficacy in high-stake graph applications \cite{Li2022OutOfDistributionGO}.
Numerous methods have been proposed to tackle distribution shifts for Euclidean data \cite{Zhuang2020ACS,liang2023comprehensive,fang2022source}.
However, applying these methods to graphs is restricted, as the interconnected entities on graphs violate the IID assumption underlying traditional machine learning methods.
Moreover, the complex graph shift types present new challenges.
These shifts could happen in different modalities including features, structures, and labels, and can be manifested in various forms such as variations in graph sizes, subgraph densities, and homophily \cite{chen2022learning}.
Given these obstacles,
increasing research efforts have been put into improving the reliability of graph machine learning against distribution shifts, ranging from graph OOD generalization~\cite{Li2022OutOfDistributionGO,chen2022learning} to graph OOD adaptation~\cite{Zhu2021ShiftRobustGO,Liu2023StructuralRI}.



Compared to graph OOD generalization, which assumes the model has no access to target data and aims to achieve satisfactory generalization performance on any unseen distribution, graph OOD adaptation takes a step further by efficiently incorporating information from the target distribution. 
With the goal of training or tuning a model to perform well under the specific target distribution, graph OOD adaptation methods excel in scenarios where integrating information from the partially observable target data is crucial,
such as transferring knowledge from the well-labeled air transport network in one region to the unlabeled air transport network in another region~\cite{Zhu2020TransferLO}, or dealing with distribution discrepancies in time-evolving citation networks~\cite{Zhu2022ShiftRobustNC}.
While several surveys have extensively investigated graph OOD generalization and its closely related techniques \cite{Li2022OutOfDistributionGO,xia2022survey},
a systematic review of graph OOD adaptation has been overlooked, despite the significance and fast growth of the area.


With recent progress on graph OOD adaptation, an up-to-date and forward-looking review of this critical problem is urgently needed. 
In this survey, we provide, to our best knowledge, the first formal and systematic review of the literature on graph OOD adaptation.
We start by formally formulating the problems and discussing different graph distribution shift types in graph machine learning.
Afterward, a new taxonomy for graph OOD adaptation is proposed, classifying existing methods into two categories based on the model learning scenario: (1) training-time graph OOD adaptation, where the distribution adaptation happens during model training on both source and target distributions \cite{You2023GraphDA,Zhu2023ExplainingAA}, and (2) test-time graph OOD adaptation, where the adaptation is performed based on a model pre-trained on the source distribution \cite{Jin2022EmpoweringGR,Zhu2023GraphControlAC}.
For each of the problems, we further categorize the existing methods as model-centric approaches and data-centric approaches. Within each subline of research, we elaborate on the detailed techniques for mitigating distribution shifts on graphs.
Based on the current progress on graph OOD adaptation, we also point out several promising research directions in this evolving field.

\section{Graph Out-Of-Distribution Adaptation}
\label{sec:Pre}
\subsection{Problem Definition}
Let $\mathcal{V} = \{i | 1\leq i\leq N\}$ denote the node set of a graph $\mathcal{G} = (\textbf{A}, \textbf{X})$, where $\textbf{A} = \{a_{uv} | u,v \in \mathcal{V} \}$ is the adjacency matrix and $\textbf{X} = \{x_v | v \in \mathcal{V} \}  $ is the node feature matrix. 
Denote a graph model characterized by parameters $\theta$ as $\varphi_{\theta}$.
For node-level or edge-level tasks, we adopt a local view and fragment the graph as a set of k-hop subgraphs of the focal node or edge to accommodate the non-iid nature of graph entities, adhering to previous works \cite{wu2022handling,Zhu2020TransferLO}.
Consequently,
for node-level tasks, 
the graph model can be written as $\varphi_{\theta}(\cdot): \mathcal{G}_v \rightarrow y_v$, where $\mathcal{G}_v$ and $y_v$ denote the k-hop subgraph  for learning the node representations and the label of the node $v$;
for edge-level tasks, the graph model can be written as $\varphi_{\theta}(\cdot): \mathcal{G}_{u,v} \rightarrow y_{u,v}$, where $\mathcal{G}_{u,v}$ and $y_{u,v}$ represent the k-hop subgraph for learning the edge representation and the label of the edge $(u,v)$;
and for graph-level tasks, the model can be written as $\varphi_{\theta}(\cdot): \mathcal{G} \rightarrow y_{\mathcal{G}}$, where $y_{\mathcal{G}}$
is the label of the entire graph. 
As a whole, the model can be denoted as 
 $\varphi_{\theta}(\cdot): ( \textbf{A}, \textbf{X} ) \rightarrow \textbf{Y}$, where $\textbf{Y}$ represents the label (matrix) of the graph. 
Without loss of generality, we focus on node-level tasks in the following problem definition, while this can naturally be extended to edge-level and graph-level tasks.

\begin{problem}
\textbf{Training-time Graph OOD Adaptation}:
Let $\mathcal{V}_{S}$ denote the node set for instances from the source distribution $\mathcal{P}_{S}(\mathcal{G}_v, y_v)$,$\mathcal{V}_{T}^\mathcal{L}$ and $\mathcal{V}_{T}^\mathcal{U}$ denote the node set for labeled and unlabeled instances from the target distribution $\mathcal{P}_{T}(\mathcal{G}_v, y_v)$.
Given source instances ~$\mathcal{D}_S=\{(\mathcal{G}_v, y_v)\}_{v \in \mathcal{V}_S}$ and target instances 
~$\mathcal{D}_T=\{ \mathcal{G}_v\}_{v \in  \mathcal{V}_{T}^\mathcal{U}} \cup  
\{(\mathcal{G}_v, y_v)\}_{v \in \mathcal{V}_{T}^\mathcal{L}}$.
Under the assumption that there exist distribution shifts between source and target ~$\mathcal{P}_{S}(\mathcal{G}_v, y_v) \neq \mathcal{P}_{T}(\mathcal{G}_v, y_v)$,
the goal of training-time graph OOD adaptation is to learn an optimal model $\varphi_{{\theta}^*}$ based on the given instances,  such that 
\begin{equation}
\label{eq:problem}
    \varphi_{{\theta}^*} = \arg \min _{\theta} \mathbb{E}_{\mathcal{P}_{T}(\mathcal{G}_v, y_v)} [l(\varphi_{\theta}(\mathcal{G}_v), y_v)]. 
\end{equation}
\end{problem}

\begin{problem}
\textbf{(Test-time Graph OOD Adaptation)}:
Given a model pre-trained on source instances $\varphi_{\theta^\prime}$, and target instances ~$\mathcal{D}_T=\{ \mathcal{G}_v\}_{v \in  \mathcal{V}_{T}^\mathcal{U}} \cup  
\{(\mathcal{G}_v, y_v)\}_{v \in \mathcal{V}_{T}^\mathcal{L}}$, the goal of test-time graph OOD adaptation is to adapt
the pre-trained model so that it achieves Equation \ref{eq:problem} under the condition that $\mathcal{P}_{S}(\mathcal{G}_v, y_v) \neq \mathcal{P}_{T}(\mathcal{G}_v, y_v)$.  
\end{problem}
\noindent

\begin{figure}[h]
\centering
\includegraphics[width=0.50\textwidth, height=0.3\textwidth]{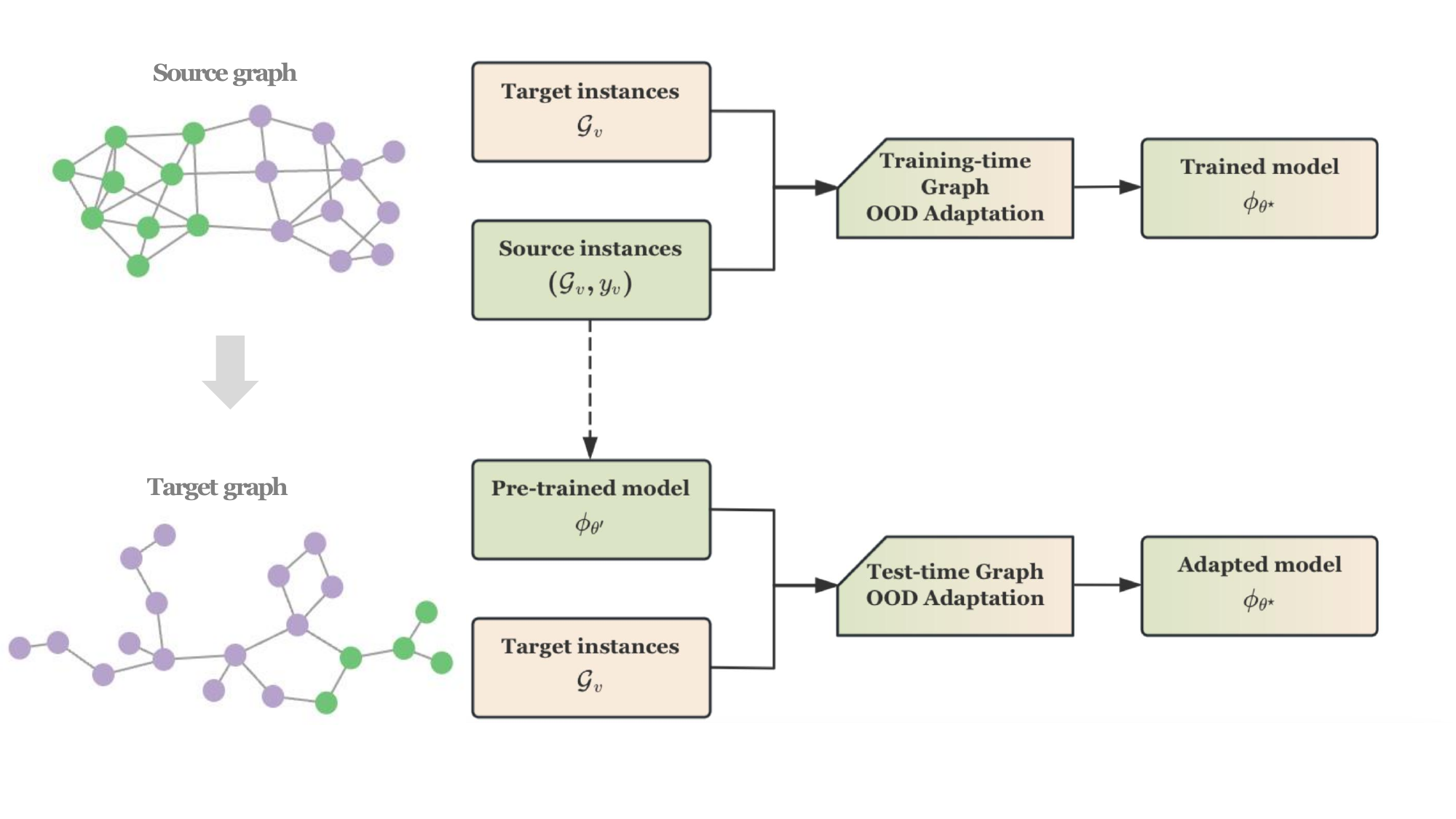}
\caption{An illustration of training-time graph OOD adaptation and test-time graph OOD adaptation.}
\label{Fig:concepts}
\end{figure}

Further, we call the problem unsupervised if $\mathcal{V}_{T}^\mathcal{L} = \varnothing$, namely, none of the target instances are labeled. Otherwise, when a proportion of target instances are labeled, we call it a semi-supervised problem.
An illustration of training-time graph OOD adaptation and test-time graph OOD adaptation can be found in Figure~\ref{Fig:concepts}.

\noindent
\textbf{Graph Distribution Shift Types.} 
In traditional machine learning, several studies have discussed and defined various types of distribution shifts \cite{moreno2012unifying}, \cite{kull2014patterns}, of which the most widely-used concepts are \textit{covariate shifts} (shifts in $\mathcal{P}(\textbf{X})$) and \textit{concept shifts} (shifts in $\mathcal{P}(\textbf{X}|\textbf{Y})$ or $\mathcal{P}(\textbf{Y}|\textbf{X})$).
These concepts can naturally be extended to graph setting by replacing feature inputs $\textbf{X}$ with the graph inputs $\mathcal{G}=(\textbf{A}, \textbf{X})$.
\begin{itemize}[wide=0pt]
    \item \textit{Covariate Shifts.} Covariate shifts on graphs emphasize the changes in graph inputs  $\mathcal{P}(\textbf{A}, \textbf{X})$, which can be further decomposed and interpreted as structure shifts, size shifts, and feature shifts~\cite{Li2022OutOfDistributionGO}.
    \item  \textit{Concept Shifts.} Concept shifts on graphs highlight the shifts in the relationship between graph inputs and labels 
    $\mathcal{P}(\textbf{Y}|\textbf{A}, \textbf{X})$ or $\mathcal{P}(\textbf{A}, \textbf{X}|\textbf{Y})$. 
    The concept shifts can be further decomposed to reveal more specific graph distribution shift types, such as the recently proposed conditional structure shift $\mathcal{P}(\textbf{A}|\textbf{Y})$ \cite{Liu2023StructuralRI}.
\end{itemize}

Additionally, the usage of these concepts often extends beyond the input space to the latent representation space $\mathcal{H}$, with  
covariate shifts describing the distribution shifts in latent representations $\mathcal{P}(\textbf{H})$, and concept shifts describing the changes in $\mathcal{P}(\textbf{H}|\textbf{Y})$ or $\mathcal{P}(\textbf{Y}|\textbf{H})$.

\begin{figure}[t!] 
\centering
\includegraphics[width=0.5\textwidth, height=0.30\textwidth]{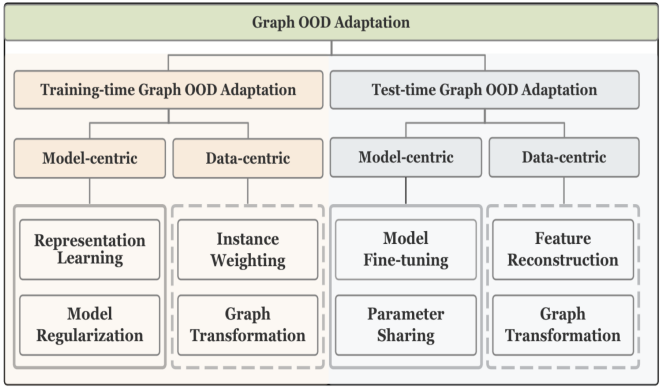}
\caption{Overview of our proposed taxonomy}
\label{Fig:tax}
\end{figure}

\subsection{Discussion on Related Topics}
Several topics are closely related to graph OOD adaptation, including: graph transfer learning, graph domain adaptation, graph OOD generalization, and fair and debiased graph learning. 
These topics, sharing similar goals with graph OOD adaptation but exhibiting nuanced differences, are further discussed in this subsection. 

\noindent
\textbf{Graph Transfer Learning.} 
In comparison to graph OOD adaptation, graph transfer learning encompasses a broader scope and does not specifically focus on addressing distribution shifts.
It involves the transfer of knowledge across distribution changes and across distinct tasks, leveraging knowledge acquired from one graph-related domain or task to enhance performance in another context. 

\smallskip
\noindent
\textbf{Graph Domain Adaptation.}
Following traditional domain adaptation, graph domain adaptation methods typically rely on the covariate shift assumption about an invariant relationship between graph inputs and labels, represented as $P_S(\textbf{Y}|\textbf{A}, \textbf{X})=P_T(\textbf{Y}|\textbf{A}, \textbf{X})$.
The objective is then to address distribution shifts in the input graph space across domains $P_S(\textbf{A}, \textbf{X}) \neq P_T(\textbf{A}, \textbf{X})$. 
In contrast to graph domain adaptation, the focus of this survey -- graph OOD adaptation is more general and comprehensive, involving distribution shifts that go beyond the covariate shift assumption.

\smallskip
\noindent
\textbf{Graph OOD Generalization.} 
Graph OOD generalization and graph OOD adaptation pursue analogous objectives in developing models capable of handling OOD target data.
In graph OOD generalization, where target data is usually assumed to be inaccessible, the primary focus is on training the model for broad generalizability, ensuring the model's effectiveness on test graphs from any potential unseen distribution.
In contrast, graph OOD adaptation fully leverages the observed target data and aims to adapt the model to the specific target distribution.

\smallskip
\noindent
\textbf{Fair and Debiased Graph Learning.}
To promote fairness among sensitive groups or mitigate GNN-induced bias issues, an ideal graph model should satisfy the following condition: 
$\hat{\mathcal{P}}(\textbf{Y}|\textbf{A},  \textbf{X}, \textbf{S}=0)=\hat{\mathcal{P}}(\textbf{Y}|\textbf{A},  \textbf{X}, \textbf{S}=1)$, where $\hat{\mathcal{P}}$ represents the predicted probability of the model, and $\textbf{S}$ indicates the latent group related to sensitive attributes such as gender \cite{kose2022fair}, or bias-related structural information such as the node degree \cite{ju2023graphpatcher}. 
Although both share the goal of alleviating distribution discrepancies, fair and debiased graph learning strives to mitigate the discrepancies in the estimated posterior distributions between different groups to ensure fairness.
On the other hand, graph OOD adaptation focuses on handling discrepancies in the population distribution between training and test to improve model performance.


\subsection{Taxonomy}
From previous problem definitions, training-time and test-time graph OOD adaptation significantly differ in model learning scenarios, with training-time adaptation starting from scratch while test-time adaptation starts from a pre-trained model.
Consequently, in the following two sections, we first categorize existing methods into training-time graph OOD adaptation and test-time graph OOD adaptation.
Within each section, we follow related surveys \cite{Zhuang2020ACS} \cite{yu2023comprehensive} and further classify methods into model-centric and data-centric approaches.
Model-centric approaches center on the learning process or the design of the graph model, while data-centric approaches emphasize the manipulation of input graphs, such as adjusting input instances or transforming graph structure or features.
Our taxonomy is shown in Figure \ref{Fig:tax}.


\section{Training-Time Graph OOD Adaptation}
\label{sec:TrTA}
Generally, training-time graph OOD adaptation serves three primary objectives in different scenarios.
\begin{itemize}[wide=0pt]
    \item \textit{Observation Bias Correction.} For semi-supervised classification within a single graph, distribution shifts between training and test instances may arise from observation bias related to latent subpopulation \cite{Bi2023PredictingTS}, or the time-evolving nature of graphs \cite{Zhu2022ShiftRobustNC}. 
    Mitigating distribution shifts in this setting may enhance the model's performance on test instances.
    \item \textit{Cross-graph Knowledge Transfer.}
    In order to transfer knowledge from well-labeled graphs to graphs with limited labels, it is crucial to properly handle distribution shifts between graphs since distinct graphs typically exhibit varied data distributions.
    \item \textit{Negative Augmentation Mitigation.}
    Graph data augmentation,  which utilizes the augmented data as additional training data, is commonly used for improving model generalization or alleviating label scarcity issues. However, overly severe distribution shifts between the original and augmented data may lead to the negative augmentation problem \cite{wu2022knowledge}, \cite{liu2022confidence}. 
    Therefore, controlling distribution shifts is essential to avoid inferior model performance and fully exploit the benefits of augmented data. 
\end{itemize}
In this section, we discuss existing training-time graph OOD adaptation methods, highlighting the techniques for mitigating distribution shifts behind these methods.
Additional information such as the task, objective, and supervision, can be found in Table \ref{tab:1}.

\subsection{Model-Centric Approaches}
In this subsection, we introduce model-centric approaches for training-time graph OOD adaptation.
These approaches can be further categorized into \textit{distributionally aligned representation learning} which aims at learning aligned representations, and \textit{model regularization} which focuses on achieving effective knowledge transfer through model regularization.

\smallskip 
\noindent 
\textbf{Distributionally Aligned Representation Learning.} 
Generally, the deep graph model $\varphi$ can be decomposed as $f \circ g$, where $g (\cdot): (\textbf{A}, \textbf{X}) \rightarrow \textbf{H}$ is a representation learner mapping the graph inputs to latent representations $\textbf{H}$, and $f(\cdot)$ is a classifier in the latent space.  
Existing literature on learning aligned representations can be further divided into
\textit{domain-invariant representation learning} and \textit{concept-shift aware representation learning}.

\begin{itemize}[wide=0pt]
\item 
\textit{Domain-invariant Representation Learning} 
is frequently employed for domain adaptation under covariate assumption, in which an invariant relationship between latent representations and labels $P_S(\textbf{Y}|\textbf{H})=P_T(\textbf{Y}|\textbf{H})$ is assumed.
Inspired by the theoretical generalization bound \cite{ben2006analysis}, domain-invariant representation learning methods aim to train a representation learner $g(\cdot)$ such that the discrepancies between the induced marginal source distribution $\mathcal{P}_S(\textbf{H})$ and target distribution $\mathcal{P}_T(\textbf{H})$ can be reduced, and at the same time, to find a classifier $f(\cdot)$ in the latent space that achieves small empirical source risk. To achieve these two goals, the loss function for domain-invariant representation learning is usually formulated as:

\begin{equation}
\label{eq:1}
\min_{f,g} \mathbb{E}_{\textbf{A, X, Y}} [l(f ( g (\textbf{A}, \textbf{X}), \textbf{Y}) ) ] + l_{reg},
\end{equation}
where $l_{reg}$ denotes a regularization term that
facilitates the alignment of the induced marginal distribution $\mathcal{P}(\textbf{H})$. 
Three strategies are mainly adopted:
\textit{explicit distance minimization}, \textit{adversarial training}, and \textit{disentangled learning}.

    \begin{itemize}[wide]
    \item \textit{(1) Explicit Distance Minimization}
    directly employs the distance between marginal distributions as the regularization term in Equation \ref{eq:1}.
    Methods vary in terms of the choice of distance metric and the specific representations they aim to align.  SR-GNN \cite{Zhu2021ShiftRobustGO} considers central moment discrepancy as regularization and aligns distribution discrepancies in the final layer of traditional GCN.
    CDNE \cite{Shen2019NetworkTN}, GraphAE \cite{Guo2023LearningAN} and GRADE \cite{Wu2022NonIIDTL} target at minimizing the statistical discrepancies between source and target across all latent layers, with the regularization term as a summation of distribution distances of different layers.
    Specifically, CDNE uses marginal maximum mean discrepancy and class-conditional marginal maximum mean discrepancy,
    GraphAE considers the multiple kernel variant of maximum mean discrepancy as the distance metric, and
    GRADE defines and utilizes subtree discrepancy. 
    JHGDA \cite{Shi2023ImprovingGD} relies on a hierarchical pooling module to extract network hierarchies and
    minimizes statistical discrepancies in hierarchical representations via the exponential form of marginal and class-conditional maximum mean discrepancy.
    For non-trainable representations, for instance, the latent embeddings in SimpleGCN \cite{wu2019simplifying},
    SR-GNN \cite{Zhu2021ShiftRobustGO} employs an instance weighting technique in which the learnable weight parameters are optimized through kernel mean matching to alleviate the distribution discrepancies.
    \item \textit{(2) Adversarial Learning} 
    aligns the representations by training the representation learner $g(\cdot)$ to generate embeddings that confuse the domain discriminator $f_d(\cdot)$. 
    Correspondingly, the regularization term in Equation \ref{eq:1} is usually framed as a minimax game between $g(\cdot)$ and $f_d(\cdot)$ as:
    \[\min_{f_d} \max_{g} l(f_d(g(\textbf{A}, \textbf{X})), \textbf{Y}_d),\]
    where $Y_d$ denotes the domain label, and $l$ can be chosen as a negative distance loss \cite{Dai2019GraphTL}, or a domain classification loss \cite{Wu2019DomainAdversarialGN,Wu2020UnsupervisedDA,Shen2020AdversarialDN,Guo2023LearningAN,Qiao2023SemisupervisedDA}.
    Instead of framing it as a minimax problem, authors \cite{Zhang2019DANEDA} explore using two symmetric and adversarial losses to train the representation learner and domain classifier, aiming to achieve bi-directional transfer.
    Typically, adversarial alignment takes place in the final hidden layer, with the exception being GraphAE \cite{Guo2023LearningAN}, which aligns representations in all hidden layers.
    In addition, it is noteworthy that SGDA \cite{Qiao2023SemisupervisedDA} also takes the label scarcity issue of the source graph into account by employing a weighted self-supervised pseudo-labeling loss.
    \item \textit{(3) Disentangled Learning}
     decomposes representations into several understandable components, with one of them being domain-invariant and related to semantic classification. The loss function for disentangled representation learning takes the form:
    \[\min_{f, g_s} \mathbb{E}_{\textbf{A, X, Y}} [l(f ( g_s (\textbf{A}, \textbf{X}), \textbf{Y}))] +  \min_{g_s, g_o} (l_{reg} + l_{recon} + l_{add}),\]
    where
    $g_s$ 
    denotes the representation learner for acquiring domain-invariant classification-related information, 
    $g_o$ denotes representation learner(s) for other components excluding $g_s$,
    $l_{reg}$ represents a regularization term for enhancing the separation between different components, 
    and 
    $l_{recon}$ denotes a reconstruction loss aiming to recover the original graph structure from the concatenated representation, thereby preventing information loss.
    Additional terms $l_{add}$ are introduced to facilitate the learning of disentangled representations, enabling specific components to exhibit desired characteristics.
    In ASN \cite{Zhang2021AdversarialSN}, the representation is decomposed into a domain-private part and a domain-invariant classification-related part. A domain adversarial loss is additionally added to facilitate the learning of invariant representations.
    Analogous to DIVA \cite{Ilse2019DIVADI}, DGDA \cite{Cai2021GraphDA} assumes that the graph generation process is controlled independently by domain-invariant semantic latent variables, domain latent variables, and random latent variables. 
    To learn representations with desired characteristics, domain classification loss and noise reconstruction loss are considered as the additional losses.
\end{itemize}
\item \textit{Concept-shift Aware Representation Learning} extends beyond the scope of learning domain-invariant representations and takes the change of label function across domains into consideration. 
Domain-invariant representation learning that minimizes the empirical source risk and the marginal distribution discrepancy inherently relies on the covariate shift assumption about invariant $\mathcal{P}(\textbf{Y}|\textbf{H})$, leading to the inestimable term in the generalization bound equal to zero \cite{ben2006analysis}.
However, as illustrated in \cite{zhao2019learning},
when there exist concept shifts in $\mathcal{P}(\textbf{Y}|\textbf{X})$ or $\mathcal{P}(\textbf{Y}|\textbf{H})$, namely, the label function changes, the inestimable adaptability term in the upper bound \cite{ben2006analysis} may be large and the performance of domain-invariant representation learning methods on target is no longer guaranteed.
A similar upper bound and an illustrative example are also provided in~\cite{Liu2023StructuralRI}, illustrating the insufficiency of domain-invariant representation learning.

To further accommodate the change in label function, 
SRNC \cite{Zhu2022ShiftRobustNC} leverages graph homophily, incorporating a shift-robust classification GNN module and an unsupervised clustering GNN module to alleviate the distribution shifts in joint distribution $\mathcal{P}(\textbf{H}, \textbf{Y})$.
Notably, SRNC is also capable of handling the open-set setting where new classes emerge in the test data.
In StruRW \cite{Liu2023StructuralRI}, authors identify and then mitigate the conditional structure shifts $\mathcal{P}(\textbf{A}|\textbf{Y})$.
They adaptively adjust the weights of edges in the source graph during training to align the distribution of a source node's neighborhood with that of target nodes from the same pseudo-class under the contextual stochastic block model. 
However, how to further align the shifts in $\mathcal{P}(\textbf{Y})$ and $\mathcal{P}(\textbf{X}|\textbf{Y})$ is left for future studies.
Moreover, authors \cite{Zhu2023ExplainingAA} demonstrate, under contextual stochastic block model, that the conditional shifts in latent space $\mathcal{P}(\textbf{Y}|\textbf{H})$ can be exacerbated by both graph heterophily and the graph convolution in GCN compared with the conditional shifts in input feature space $\mathcal{P}(\textbf{Y}|\textbf{X})$. 
Hence, they introduce GCONDA that explicitly matches the distribution of $\mathcal{P}(\textbf{Y}|\textbf{H})$ across domains via Wasserstein distance regularization, and additionally, they also propose GCONDA++ that jointly minimizes the discrepancy in $\mathcal{P}(\textbf{Y}|\textbf{H})$ and $\mathcal{P}(\textbf{H})$. 
\end{itemize}

\smallskip 
\noindent
\textbf{Model Regularization.}
Instead of focusing on the process of learning aligned representations, some other methods achieve effective knowledge transfer under distribution shifts through model regularization. 
Building on the derived GNN-based generalization bound, authors \cite{You2023GraphDA}  propose SSReg and MFRReg, which regularize the spectral properties of GNN to enhance transferability.
They also extend their theoretical results to the semi-supervised setting with the challenging distribution shifts in $\mathcal{P}(\textbf{Y}|\textbf{A}, \textbf{X})$.
Both KDGA~\cite{wu2022knowledge} and KTGNN \cite{Bi2023PredictingTS} employ knowledge distillation, regularizing the Kullback–Leibler divergence between the outputs of teacher and student models.
Particularly, KDGA aims to mitigate the negative augmentation problem by distilling the knowledge of a teacher model trained on augmented graphs to a partially parameter-shared student model on the original graph.
KTGNN, on the other hand, considers the semi-supervised node classification problem for VS-Graph, in which vocal nodes are regarded as the source and silent nodes with incomplete features are regarded as the target. 
They apply a domain-adapted feature completion module and domain-adapted message-passing mechanism to learn representations that capture domain differences.
Then, the source classifier and target classifier are respectively constructed, and the knowledge of both source and target classifiers is distilled into the student transferable classifier through the KL regularization.

\begin{table*}[ht]
\centering
\scalebox{0.61}{
\begin{tabular}{ccccccc}
\hline
\rowcolor{Gray}  &  &  &  &  &  &  \\
\rowcolor{Gray}  
\multirow{-2}{*}{\textbf{Category}} & \multirow{-2}{*}{\textbf{Name}} & \multirow{-2}{*}{\textbf{Reference}} & \multirow{-2}{*}{\textbf{Task Level}} & \multirow{-2}{*}{\textbf{Distribution Shift}} & \multirow{-2}{*}{\textbf{Objective}} & \multirow{-2}{*}{\textbf{Supervision}}  \\ \midrule
\multirow{17}{*}{\makecell{Domain-invariant\\Representation Learning}} 
& DAGNN & \cite{Wu2019DomainAdversarialGN} & graph & $\mathcal{P}(\textbf{A}, \textbf{X})$ & Cross-graph transfer & unsupervised \\  \cmidrule{2-7}
& DANE & \cite{Zhang2019DANEDA} & node & $\mathcal{P}(\textbf{A}, \textbf{X})$ & Cross-graph transfer & unsupervised \\  \cmidrule{2-7}
& CDNE & \cite{Shen2019NetworkTN} & node & $\mathcal{P}(\textbf{A}, \textbf{X}, \textbf{Y})$ & Cross-graph transfer & semi-supervised \\  \cmidrule{2-7}
& ACDNE & \cite{Shen2020AdversarialDN} & node & $\mathcal{P}(\textbf{A}, \textbf{X})$ & Cross-graph transfer & unsupervised \\  \cmidrule{2-7}
& UDA-GCN & \cite{Wu2020UnsupervisedDA} & node & $\mathcal{P}(\textbf{A}, \textbf{X})$ & Cross-graph transfer & unsupervised \\  \cmidrule{2-7}
 & DGDA & \cite{Cai2021GraphDA} & graph & $\mathcal{P}(\textbf{A}, \textbf{X})$ & Cross-graph transfer & unsupervised \\ \cmidrule{2-7}
& SR-GNN & \cite{Zhu2021ShiftRobustGO} & node & $\mathcal{P}(\textbf{A}, \textbf{X})$ & Observation bias correction & unsupervised \\ \cmidrule{2-7}
& ASN & \cite{Zhang2021AdversarialSN} & node & $\mathcal{P}(\textbf{A}, \textbf{X})$ & Cross-graph transfer & unsupervised \\  \cmidrule{2-7}
 & AdaGCN & \cite{Dai2019GraphTL} & node &$\mathcal{P}(\textbf{A}, \textbf{X})$   $ /\ $ $\mathcal{P}(\textbf{A}, \textbf{X}, \textbf{Y})$ & Cross-graph transfer & un/semi-supervised \\  \cmidrule{2-7}
 & GraphAE & \cite{Guo2023LearningAN} & node & $\mathcal{P}(\textbf{A}, \textbf{X})$& Cross-graph transfer & unsupervised \\  \cmidrule{2-7}
 & GRADE & \cite{Wu2022NonIIDTL} & node / edge & $\mathcal{P}(\textbf{A}, \textbf{X})$ & Cross-graph transfer & unsupervised \\  \cmidrule{2-7}
 & JHGDA & \cite{Shi2023ImprovingGD} & node & $\mathcal{P}(\textbf{A}, \textbf{X})$ & Cross-graph transfer & unsupervised \\  \cmidrule{2-7}
 & SGDA & \cite{Qiao2023SemisupervisedDA} & node & $\mathcal{P}(\textbf{A}, \textbf{X})$ & Cross-graph transfer & unsupervised \\  
  \midrule
\multirow{3}{*}{\makecell{Concept-shift Aware\\ 
Representation Learning}} & SRNC & \cite{Zhu2022ShiftRobustNC} & node & $\mathcal{P}(\textbf{A}, \textbf{X},\textbf{Y})$ & Cross-graph transfer / Observation bias correction & unsupervised \\ \cmidrule{2-7}
 & StruRW & \cite{Liu2023StructuralRI} & node & $\mathcal{P}(\textbf{A}|\textbf{Y})$ & Cross-graph transfer & unsupervised \\ \cmidrule{2-7}
& GCONDA++ & \cite{Zhu2023ExplainingAA} & node / graph & $\mathcal{P}(\textbf{A}, \textbf{X},\textbf{Y})$ & Cross-graph transfer / Observation bias correction & unsupervised \\  \midrule
\multirow{2}{*}{Model Regularization} 
 & KDGA & \cite{wu2022knowledge} & node & $\mathcal{P}(\textbf{A},\textbf{Y})$ & Negative augmentation mitigation& semi-supervised \\ \cmidrule{2-7}
 & SS/MFR-Reg & \cite{You2023GraphDA} & node / edge & $\mathcal{P}(\textbf{A}, \textbf{X})$   $ /\ $ $\mathcal{P}(\textbf{A}, \textbf{X}, \textbf{Y})$ & Cross-graph transfer & un/semi-supervised  \\ 
\cmidrule{2-7}
 & KTGNN & \cite{Bi2023PredictingTS} & node & $\mathcal{P}(\textbf{X},\textbf{Y})$ & Observation bias correction & semi-supervised \\ \midrule
\multirow{4}{*}{Instance weighting} & IW & \cite{Ye2013PredictingPA} & edge & $\mathcal{P}(\textbf{A},\textbf{Y})$ & Cross-graph transfer & semi-supervised \\ \cmidrule{2-7}
 & NES-TL & \cite{Fu2020NESTLNE} & node & $\mathcal{P}(\textbf{A}, \textbf{Y})$ & Cross-graph transfer & semi-supervised \\ \cmidrule{2-7}
 & RSS-GCN & \cite{Wu2022ReinforcedSS} & graph &$\mathcal{P}(\textbf{A}, \textbf{X})$ & Cross-graph transfer & unsupervised \\ \cmidrule{2-7}
 & DR-GST & \cite{liu2022confidence} & node & $\mathcal{P}(\textbf{A}, \textbf{X},\textbf{Y})$ & Negative augmentation mitigation& semi-supervised \\  \midrule
\multirow{2}{*}{Graph Transformation}  & FakeEdge & \cite{Dong2022FakeEdgeAD} & edge & $\mathcal{P}(\textbf{A}, \textbf{Y})$ & Observation bias correction & unsupervised \\ \cmidrule{2-7}
 & Bridged-GNN & \cite{Bi2023BridgedGNNKB} & node & $\mathcal{P}(\textbf{A}, \textbf{X},\textbf{Y})$ & Cross-graph transfer & semi-supervised \\ \cmidrule{2-7}
 & DC-GST & \cite{wang2024distribution} & node & $\mathcal{P}(\textbf{A}, \textbf{X})$ & 
Negative augmentation mitigation $\&$
Observation bias correction & unsupervised \\ 
\midrule
\end{tabular}
}
\caption{A summary of training-time graph OOD adaptation methods. `Task Level' denotes the main task level, `Distribution shift' denotes the distribution shifts the methods aim to handle, `Objective' denotes the three primary objectives discussed at the beginning of Section \ref{sec:TrTA}, and `Supervision' indicates whether a proportion of instances from target distribution are labeled (semi-supervised) or not (unsupervised).}
\label{tab:1}
\end{table*}

\subsection{Data-Centric Approaches}
\smallskip
\noindent
\textbf{Instance Weighting.} 
Instance Weighting, which assigns different weights for instances, is a commonly used data-centric technique in traditional transfer learning \cite{Zhuang2020ACS}. Similar strategies are observed in methods for training-time graph OOD adaptation.
Borrowing the idea from Adaboost and TrAda,  authors \cite{Ye2013PredictingPA} employ the instance weighting technique for the edge sign prediction task.
The edge weights are adjusted in each iteration, with reduced weights assigned to misclassified dissimilar source instances to mitigate distribution shifts across graphs. 
Authors \cite{liu2022confidence} recognize that
the distribution shifts between the original data and the augmented data with pseudo-labels may impede the effectiveness of self-training. 
To mitigate the gap between the original distribution and the shifted distribution, 
they assign weights to augmented node instances based on information gain, paying more attention to nodes with high information gain rather than those with high confidence.
Both RSS-GCN and NES-TL consider the multi-source transfer problem, where multiple graphs are available as source. 
Since the source graphs may not be equally important for predictions on the target graph and some of them may be of poor quality, a weighting technique is employed to effectively combine the available source graphs.
NES-TL \cite{Fu2020NESTLNE} proposes the NES index to quantitatively measure the structural similarity between two graphs, and use the NES-based scores as weights to ensemble weak classifiers trained on instances from each source graph and labeled target instances.
RSS-GCN \cite{Wu2022ReinforcedSS} utilizes reinforcement learning to select high-quality source graphs for multi-source transfer, aiming to minimize the distribution divergence between selected source graphs and target graphs. Such sample selection strategy can be considered as a special binary instance weighting.

\smallskip
\noindent
\textbf{Graph Transformation.} 
Several authors have delved into the exploration of leveraging graph transformation strategies to alleviate the distribution shifts, through adding or removing edges. Authors \cite{Dong2022FakeEdgeAD} find that the dataset shift challenge in edge prediction arises from the presence of links in training and the absence of link observations in testing.
To tackle this challenge, they propose FakeEdge, a subgraph-based link prediction framework that intentionally adds or removes the focal link within the subgraph. This adjustment decouples the dual role of edges as elements in representation learning and as labels of links in link prediction, thereby ensuring that the subgraph is consistent across training and testing.
Additionally, authors \cite{Bi2023BridgedGNNKB} 
reconsider the domain-level knowledge transfer problem as learning sample-wise knowledge-enhanced posterior distribution. 
They first learn the similarities of samples from both source and target graphs and build bridges between each sample and its similar samples containing valuable knowledge for prediction.
A GNN model is then employed to transfer knowledge across source and target samples on the constructed bridged-graph.
More recently, a novel framework called DC-GST \cite{wang2024distribution} has been introduced to bridge the distribution shifts between augmented training instances and test instances in self-training, which incorporates a distribution-shift-aware edge predictor to improve the model's generalizability of assigning pseudo-labels.
Furthermore, they employ the distribution consistency criterion and neighborhood entropy reduction criterion for the selection of pseudo-labeled nodes.
In doing so, they aim to identify nodes that are not only informative but also effective in mitigating the distribution discrepancy between source and target.


\section{Test-Time Graph OOD Adaptation}
\label{sec:TTA}

\smallskip 
In this section, we concentrate on graph OOD adaptation during test time. 
In training-time graph OOD adaptation, both source and target instances need to be observed simultaneously. 
However, this may be unrealistic in various graph-related applications.
For instance, in social networks,
source data is typically confidential and inaccessible due to privacy protection purposes and data leakage concerns. 
Additionally, storing the complete source data on resource-limited devices may also be impractical.
In contrast to training-time adaptation, test-time adaptation is not restricted by the availability of labeled source data and aims to adapt a pre-trained model to perform effectively on the target data.
This form of adaptation, also known as source-free adaptation, plays a crucial role in scenarios where access to source data is restricted. 
\subsection{Model-Centric Approaches}
\smallskip 
\noindent
\textbf{Model Fine-tuning.} 
Fine-tuning is a widely used approach to address graph distribution shifts during test time. However, effectively leveraging information from the pre-trained model presents challenges, generally in two scenarios. 
In the first scenario, the model is pre-trained to encode more transferable and generalizable structural information, and then task-related information and domain-specific node attributes are added during fine-tuning. 
Consequently, this scenario requires target labels, which are often very limited, and thus may lead to the overfitting problem. 
To tackle this challenge, authors \cite{Zhu2023GraphControlAC} propose GraphControl that incorporates target data as conditional inputs inspired by the success of ControlNet \cite{Zhang2023AddingCC}. 
The structural information is fed into a frozen pre-trained model, while a kernel matrix built on node features is fed into the trainable copy. 
The two components are connected through zero MLPs with gradually expanding parameters, aiming to prevent the harmful impact of noise in target node features while gradually integrating downstream information into the pre-trained model.
In the second scenario, task-related information is encoded into the pre-trained model, and subsequently, an unsupervised fine-tuning procedure is applied. 
Yet, when tuning on the unsupervised task, the model may lose the discriminatory power related to the main task, or learn irrelevant information. 
In SOGA \cite{Mao2021SourceFU}, authors utilize a loss that maximizes the mutual information between the inputs and outputs of the model to enhance the discriminatory power.
GT3 \cite{Wang2022TestTimeTF} avoids overfitting to the downstream self-supervised task by adding regularization constraints between training and test output embeddings, enforcing their statistical similarity and avoiding substantial fluctuations.
Furthermore,  GAPGC \cite{Chen2022GraphTTATT} aims to address the over-confidence bias and the risk of capturing redundant information through the use of an adversarial pseudo-group contrast strategy. From the information bottleneck perspective, GAPGC provides a lower bound guarantee of the information relevant to the main task.
Some relevant studies, such as GTOT-Tuning \cite{Zhang2022FineTuningGN}, which concentrates on transferring knowledge across tasks on the same graph during test time, also provide valuable insights and may have the potential for dealing with distribution shifts.

\smallskip 
\noindent
\textbf{Parameter Sharing.} 
Parameter sharing is a model design strategy that involves constructing models with both domain-shared and domain-specific parameters.
In these approaches, domain-shared parameters are directly employed without retraining, and adjustments are only made to domain-specific parameters during test-time.
GraphGLOW \cite{Zhao2023GraphGLOWUA} 
is designed to integrate a shared graph structure learner and dataset-specific GNN heads for classification tasks in the cross-graph transfer setting. 
In testing, only the data-specific GNN is updated while the structure learner from the pre-trained model is directly applied. 
GT3 \cite{Wang2022TestTimeTF} 
structures the model to include two branches: a main task (classification) branch and a self-supervised branch. The two branches share initial layers and have unique task-specific layers and parameters afterwards. 
During the training phase, all parameters are optimized using a combination of self-supervised loss and main task loss. 
In the test phase, the main task branch is utilized for prediction, in which the unique parameters of the branch remain unchanged, and the parameters in the initial layers are tuned based on the self-supervised task on the target graph. 

\subsection{Data-Centric Approaches}

\smallskip 
\noindent
\textbf{Feature Reconstruction.}
Feature reconstruction is a test-time feature manipulation strategy that addresses distribution shifts without adapting the model structure or retraining parameters.
In \cite{Ding2023FRGNNMT}, the authors introduce FRGNN for semi-supervised node classification. 
They utilize an MLP to establish a mapping between the output and input space of the pre-trained GNN. Subsequently, using the encoded one-hot class vectors as inputs, the MLP generates class representative representations. 
By substituting the features of labeled test nodes with the representative representations of the corresponding classes and spreading the updated information to other unlabeled test nodes through message passing, the graph embedding bias between test nodes and training nodes is anticipated to be mitigated.

\smallskip 
\noindent
\textbf{Graph Transformation.}
Apart from adjusting features, authors \cite{Jin2022EmpoweringGR} introduce a graph transformation framework called GTRANS to address distribution shifts during test time. 
The graph transformation is modeled as injecting perturbations on the graph structure and node features, which is subsequently optimized via a parameter-free surrogate loss. 
Theoretical analyses guiding the selection of surrogate loss functions are additionally provided by the authors. 

It is worth highlighting that instead of modifying the pre-trained model, data-centric test-time graph OOD adaptation focuses on adjusting the test data, and is especially beneficial when handling large-scale pre-trained models.

\section{Future Directions}
\label{sec:future}
\textbf{Theoretical Study.}
Future theoretical analyses could delve deeper into the feasibility and effectiveness of graph OOD adaptation, particularly in scenarios where the label function changes between training and test data.
There is also the need to develop theories and methodologies specifically tailored for graph data or graph models, taking the intricate structural information inherent in graphs into consideration.
Furthermore, it's worth exploring more diverse scenarios, such as universal domain adaptive node classification \cite{Chen2023UniversalDA}, graph size adaptation \cite{Yehudai2020FromLS}, and multi-source transfer.
Notably, several generalization bounds are derived from the graph transferability evaluation perspective 
 \cite{Ruiz2020GraphonNN}, \cite{Zhu2020TransferLO}, \cite{Chuang2022TreeMD}, and may 
assist in selecting high-quality source graphs in the multi-source transfer setting. 
However, identifying the optimal combination of source graphs with theoretical guarantees remains an open problem.

\smallskip
\noindent
\textbf{Test-time Graph OOD Adaptation.}
Test-time adaptation has garnered increasing attention in traditional machine learning, yet relatively few works have been conducted for graph settings.
The exploration and design of more graph-specific strategies remain crucial and promising.  
Besides, the rigorous theoretical analysis for test-time adaptation remains an open problem \cite{liang2023comprehensive}, and addressing this gap could potentially inspire the development of innovative graph test-time OOD adaptation methods.
Additionally, it is worth investigating whether recent advances in unsupervised test-time graph evaluation \cite{Zheng2023GNNEvaluator}, \cite{Zheng2024Online} can contribute to facilitating test-time graph adaptation algorithms.
Lastly, alleviating the computational cost of adapting a large pre-trained model, through data-centric graph transformation or graph prompt tuning \cite{Sun2023GraphPL}, also deserves more attention in the future.

\smallskip
\noindent
\textbf{Distribution Shifts on Complex Graphs.}
In contrast to the substantial efforts dedicated to addressing distribution shifts on regular graphs, studies on more complex graph types, such as spatial, temporal, spatial-temporal, heterogeneous, and dynamic graphs, have received comparatively less attention. Such complex graphs often exhibit diverse and dynamic patterns or involve entities and relationships of various types, introducing more intricate and nuanced distribution shifts. Furthermore, existing graph OOD adaptation methods have primarily been designed and evaluated on small networks, whereas these complex graph types, especially dynamic graphs, may be of large scale, highlighting the necessity for scalable and memory-efficient graph OOD adaptation methods.
The comprehensive exploration and efficient mitigation of distribution shifts on complex graph types are pivotal for enhancing the capabilities of graph machine learning in broader scenarios, such as recommendation systems, healthcare systems, and traffic forecasting.

\section{Conclusion}
In this survey, we examine existing graph OOD adaptation methods, covering two problem scenarios including both training-time graph OOD adaptation and test-time graph OOD adaptation.
Firstly, we establish problem definitions and explore different graph distribution shift types.
Then, we discuss topics related to graph OOD adaptation and explain our categorization.
Based on the proposed taxonomy, we systematically examine the techniques for mitigating distribution shifts in existing graph OOD adaptation methods.
Finally, we highlight several challenges and future directions.
We hope that this survey will help researchers better understand the current research progress in graph OOD adaptation.

\clearpage
\bibliographystyle{named}
\bibliography{0reference}

\end{document}